\documentclass[a4paper]{article}
\usepackage[dvips]{graphicx}
\usepackage{techrep}
\usepackage{mlapa}

\renewcommand{\vec}[1]{{\bf#1}}

\newcommand{\dd}{\mbox{d}}

\author{Ivo Kwee \hspace{8mm} Marcus Hutter \hspace{8mm} 
  J\"{u}rgen Schmidhuber \\
  {\normalsize IDSIA, Manno CH-6928, Switzerland.}  \\
  {\normalsize \tt \{ivo,marcus,juergen\}@idsia.ch} }
\title{Gradient-based Reinforcement Planning in Policy-Search Methods
  \footnote{ This is an extended version of the paper presented at the
    EWRL 2001 in Utrecht (The Netherlands). In this technical report,
    the derivation steps are presented with more detail, more
    footnotes, appendices and more (unfinished) ideas. } }
\date{November 2001}

\reportnumber{14-01} 
\reportaddnote{This work was supported by SNF grants 21-55409.98 and 
  2000-61847.00 }

\begin{document}
\makecover
\maketitle


\begin{abstract}
  We introduce a learning method called ``gradient-based reinforcement
  planning'' (GREP). Unlike traditional DP methods that improve their
  policy backwards in time, GREP is a gradient-based method that plans
  ahead and improves its policy \emph{before} it actually acts in the
  environment. We derive formulas for the exact policy gradient that
  maximizes the expected future reward and confirm our ideas
  with numerical experiments.
\end{abstract}

\section{Introduction}

It has been shown that \emph{planning} can dramatically improve
convergence in reinforcement learning
(RL)~\cite{Schmidhuber:90sandiego,suttonbarto}. However, most RL
methods that explicitly use planning that have been proposed are value
(or $Q$-value) based methods, such as \emph{Dyna-Q} or
\emph{prioritized sweeping}.

Recently, much attention is directed to so-called
\emph{policy-gradient} methods that improve their policy directly by
calculating the derivative of the future expected reward with respect
to the policy parameters. Gradient based methods are believed to be
more advantageous than value-function based methods in huge state
spaces and in POMDP settings. Probably the first gradient based RL
formulation is class of REINFORCE algorithms of

Williams~\cite{williams92simple}. Other more recent methods are, e.g.,
~\cite{baird98gradient,baxter99direct,sutton00policy}. Our approach of
deriving the gradient has the flavor of~\cite{ng99policy} who derive
the gradient using future state probabilities.

Our novel contribution in this paper is to combine gradient-based
learning with explicit planning.  We introduce ``gradient-based
reinforcement planning'' (GREP) that improves a policy \emph{before}
actually interacting with the environment. We derive the formulas for
the exact policy gradient and confirm our ideas in numerical
experiments. GREP learns the action probabilities of a probabilistic
policy for discrete problem. While we will illustrate GREP with a
small MDP maze, it may be used for the hidden state in POMDPs.


\section{Derivation of the policy gradient}

\subsubsection*{projection matrix}
Let us denote the discrete space of \emph{states} by $\mathcal{S} =
\{1,...,N\}$. Our belief on $\mathcal{S}$ is decribed by a probability
vector $\vec{s}$ of which each element $s_i$ represents the
probability of being in state $i$. We also define a set of actions
$\mathcal{A} = 1,...,K$. The stochastic policy $\mathcal{P}$ is
represented by a matrix $\vec{P}:N \times K$ with elements $P_{ki} =
p(a_k|s_i)$, i.e. the conditional probability of action $k$ in
state $i$.%
\footnote{For computational reasons, one often reparameterizes the
  policy using a Boltzmann distribution. Here in this paper the
  probability $p(a_k|s_i)$ is just given by $P_{ki}$ and we do not use
  reparameterization in order to keep the analysis clear.} %
Furthermore, let environment $\mathcal{E}$ be defined by transition
matrices $\vec{T}_k$ ($k=1,...,K$) with elements $T_{kji} =
p(s_j|s_i,a_k)$, i.e. the transition probability to $s_j$ in state
$s_i$ given action $k$.  

Now we define the \emph{projection matrix} $\vec{F}$ with elements
\begin{equation}
  F_{ji} = \sum_k T_{kji} \, P_{ki}.
\label{eq:projection}
\end{equation}
Important is that matrix $\vec{F}$ is \emph{not} modelling the
transition probabilities of the environment, but models the induced
transition probability using policy $\mathcal{P}$ in environment
$\mathcal{E}$. The induced transition probability $F_{ji}$ is a
weighted sum over actions $k$ of the transition probabilities
$T_{kij}$ with the policy parameters $P_{ki}$ as the weights.

\subsubsection*{Expected state occupancy}
Using the projection matrix $\vec{F}$, states $\vec{s}_{t}$ and
$\vec{s}_{t+1}$ are related as $\vec{s}_{t+1} = \vec{F} \vec{s}_{t}$
and therefore $\vec{s}_{t} = \vec{F}^t \vec{s}_0$, where $\vec{s}_0$ is
the state probability distribution at $t=0$. We can now define the
\emph{expected state occupancy} as
\begin{equation}
  \vec{z} = E[\vec{s}|\vec{s}_0] 
  = \sum_{t=0}^\infty \gamma^t \vec{s}_t
  = \sum_{t=0}^\infty (\gamma \vec{F})^t \vec{s}_0
  = (\vec{I} - \gamma \vec{F})^{-1} \vec{s}_0
\label{eq:occupancy}
\end{equation}
where $\gamma$ is a discount factor in order to keep the sum finite.  In the
last step, we have recognized the sum as the Neumann representation of
the inverse. Notice that $\vec{z}$ is a solution of the linear
equation
\begin{equation}
  (\vec{I} - \gamma \vec{F}) \vec{z} =  \vec{s}_0
\label{eq:system}
\end{equation}
which is just the familiar Bellman equation for the expected occupancy
probability $\vec{z}$. 


\subsubsection*{Expected reward function}
In \emph{reinforcement learning} (RL) the objective is to maximize
future reward. We define a reward vector $\vec{r}$ in the same domain
as $\vec{s}$. Using the expected occupancy $\vec{z}$ the future
expected reward $H$ is simply
\begin{equation}
  H = \langle\vec{r}, \vec{z} \rangle
\label{eq:rl-error}
\end{equation}
where $\langle \cdot,\cdot \rangle$ is the scalar vector product.%
\footnote{ In \emph{optimal control} (OC) we want to reach some target
  state under some optimality conditions (mostly minimum time or
  minimum energy). We denote $\vec{r}$ as our target distribution and
  denote the time-to-arrival as $t^*$. If $t^*$ were known beforehand
  then we
  \begin{equation}
    H = \frac{1}{2} ( \vec{r} - \vec{s}_{t^*} )^2
    = \frac{1}{2} ( \vec{r} - \vec{F}^{t^*} \vec{s}_0 )^2
  \end{equation}
  
  However the exact arrival time is mostly not known beforehand, the
  most we can do is to use minimize the (time-weighted) expected error
  \begin{equation}
    H = \sum_{t^*=0}^\infty \gamma^{t^*} 
    \frac{1}{2} ( \vec{r} - \vec{F}^{t^*} \vec{s}_0 )^2
    = \frac{1}{2} k \vec{r}^2 
    - \vec{r} \vec{z} + \frac{1}{2} \sum_{t^*=0}^\infty \gamma^{t^*} 
    \left( \vec{F}^{t^*} \vec{s}_0 \right)^2
    \label{eq:control-error}
  \end{equation}
  The first term on the right hand side is often not relevant because it
  is independent of $\vec{F}$. When we compare
  Eq.~\ref{eq:control-error} with Eq.~\ref{eq:rl-error}, we see that the
  OC error function has a quadratic term in $\vec{F}$ which the RL error
  lacks.
}

Because $\vec{z}$ is a solution of Eq.~\ref{eq:system} it is dependent
on $\vec{F}$ which in turn depends on policy $\vec{P}$. Given
$\vec{r}$ and $\vec{s}_0$, our task is to find the optimal $\vec{P}^*$
such that $H$ is maximized, i.e.
\begin{equation}
  \vec{P}^* = \arg \max_{\vec{P}} H .
\end{equation}

We can regard the calculation of the future expected reward as a
composition of two operators
\begin{equation}
  \mathcal{Q}: \vec{F} \mapsto \vec{z}
\end{equation}
which maps the transition matrix $\vec{F}$ to the expected occupancy
probabilities $\vec{z}$, and
\begin{equation}
  \mathcal{R}: \vec{z} \mapsto R
\end{equation}
which maps the probabilities $\vec{z}$, given a reward distribution
$\vec{r}$, to an expected reward value $R$.

\subsubsection*{Calculation of the policy gradient}
A variation $\delta \vec{z}$ in the expected occupancy can be related
to first order to a perturbation $\delta \vec{P}$ in the (stochastic)
policy. To obtain the partial derivatives $\partial \vec{z} / \partial
P_{ik}$, we differentiate Eq.~\ref{eq:system} with respect to $P_{ik}$
and obtain:
\begin{equation}
  - \gamma \frac{\partial \vec{F}}{\partial P_{ik}} \vec{z} 
  + (\vec{I} - \gamma \vec{F})\frac{\partial \vec{z}}{\partial P_{ik}} 
  =  0.
\label{eq:diff1}
\end{equation}
The right hand side of the equation is zero because we assume that
$\vec{s}_0$ is independent of $P_{ik}$.
Rearranging gives:
\begin{equation}
  \frac{\partial \vec{z} }{\partial P_{ik} } = 
  \gamma \vec{K} \frac{\partial \vec{F}}{\partial P_{ik} } \vec{z}
\label{eq:diff2}
\end{equation}
where $\vec{K} = (\vec{I} - \gamma \vec{F})^{-1}$. %

From Eq.~\ref{eq:rl-error} and Eq.~\ref{eq:diff1}, together with the
chain rule, we obtain the gradient of the RL error with respect to the
policy parameters $P_{ik}$:
\begin{equation}
  \frac{ \partial H } { \partial P_{ik} } =
  \frac{ \partial H } { \partial \vec{z} } 
  \frac{ \partial \vec{z} } { \partial P_{ik} } 
  = \left\langle \vec{r}, \gamma \vec{K} \frac{\partial \vec{F}} 
    {\partial P_{ik}} \vec{z} \right\rangle 
  = \gamma \left\langle \vec{K}^*\vec{r}, \frac{\partial \vec{F}} 
    {\partial P_{ik}} \vec{z} \right\rangle 
\label{eq:gradient}
\end{equation}
where $A^*$ means the adjoint operator of $A$ defined by $\langle u, A
w \rangle = \langle A^*u, w \rangle$. Let us define:
$\vec{q}=\vec{K}^*\vec{r}$. While $\vec{K}$ maps the initial state
$\vec{s}_0$ to the future expected state occupancy $\vec{z}$, its
adjoint, $\vec{K}^*$, maps the reward vector $\vec{r}$ back to
\emph{expected reward} $\vec{q}$. The value of $q_i$ represents the
(pointwise) expected reward in state
$s_i$ for policy $\vec{P}$.~%
\footnote{Indeed, this is a different way to define the traditional
  \emph{value-function}. Note that generally, neither $\vec{r}$ nor
  $\vec{q}$ are probabilities because their 1-norm is generally not 1.
}

Finally, differentiating Eq.~\ref{eq:projection} gives us
$\partial \vec{F} / \partial P_{ik}$. Inserting this into
Eq.~\ref{eq:gradient} yields:
\begin{equation}
  G_{ik} = \frac{ \partial H } { \partial P_{ik} } 
  \propto \, z_i \! \sum_j T_{kji} q_j.
\label{eq:gradient-p}
\end{equation}
In words, the gradient of $H$ with respect to policy parameter
$P_{ik}$ (i.e. the probability of taking action $a_k$ in state $s_i$)
is proportional to the expected occupancy $z_i$ times the weighted sum
of expected reward $q_j$ over next states $(j=1,...,N)$ weighted by
the transition probabilities $T_{kji}$.

Note that the gradient could also have been approximated using finite
differences which would need at least $1+n^2$ field
calculations.\footnote{ Finite difference approximation of the
  derivative $\partial \vec{z} / \partial T_{ij}$ involves computing
  $\vec{z}$ for $\vec{F}$ and then perturbing a single $T_{ij}$ in
  $\vec{F}$ by a tiny amount $dT$ and subsequently recomputing
  $\vec{z}'$. Then the derivative is approximated by $\partial \vec{z}
  / \partial T_{ij} \approx (\vec{z}'-\vec{z})/dT$. For a $n\times n$
  matrix $\vec{F}$, one would need to repeat this for every element
  and would require a total upto $1+n^2$ calculations of $\vec{z}$.}
The adjoint method is much more efficient and needs only \emph{two}
field calculations.


Once we have the gradient $\vec{G}$, improving policy $\vec{P}$ is now
straight forward using gradient ascent or we can also use more
sophisticated gradient-based methods such as nonlinear conjugate
gradients (as in~\cite{baxter99direct}). The optimization is nonlinear
because $\vec{z}$ and $\vec{r}$ themselves depend on the current
estimate of $\vec{P}$.

\section{Computation of the optimal policy}

We will introduce two algorithms that incorporate our ideas of
gradient-based reinforcement planning. The first algorithms describes
an off-line planning algorithm that finds the optimal policy but
assumes that the environment transition probabilities are known. The
second algorithm is an online version that could cope with unknown
environments.

\subsection{Offline GREP}

If the environment transition probabilities $T_{kji}$ are known, the
agent may improve its policy using GREP. Our offline GREP planning
algorithm consist of two steps:

\begin{enumerate} 
\item {\em Plan ahead:} Compute the policy gradient $\vec{G}$ in
  Eq.~\ref{eq:gradient} and improve current policy
  \begin{equation}
    \vec{P} \leftarrow \vec{P} + \alpha \vec{G}
    \label{eq:t-update}
  \end{equation}
  where $\alpha$ is a suitable step size parameter; for efficiency we
  can also perform a linesearch on $\alpha$. 
\item {\em Evaluate policy:} Repeat above until policy is optimal.
\end{enumerate}

Matrix $\vec{P}$ describes a probabilistic policy. We define the
\emph{maximum probable policy} (MPP) to be the deterministic policy by
taking the maximum probable action at each state. It is not obvious
that the MPP policy will converge to the global optimal solution but
we expect MPP at least to be near-optimal. 

\subsubsection*{Numerical experiments}

\begin{figure} \centering
  \includegraphics[width=0.147\textwidth]{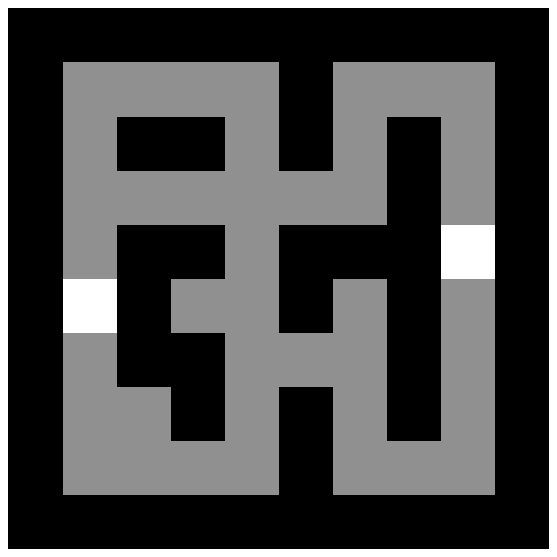}
  \includegraphics[width=0.155\textwidth]{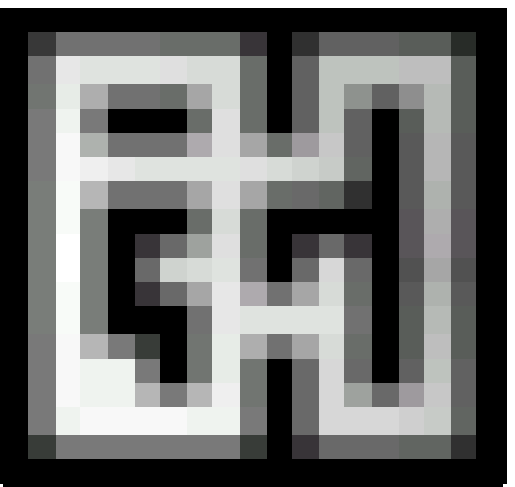}
  \includegraphics[width=0.155\textwidth]{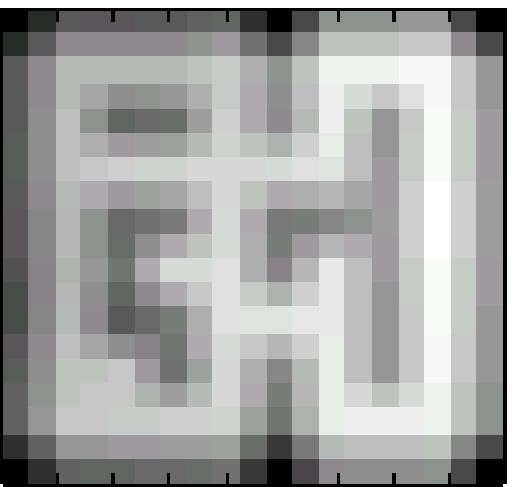}
\caption{Left: $10\times10$ toy maze with start at left and goal at right side.
  Center: plot of expected occupancy $\vec{z}$. Right: plot of
  expected reward $\vec{q}$. White corresponds to higher probability.
  [Blurring is due to visualisation only]. }
\label{fig:maze}
\end{figure}

\begin{figure*} \centering
  \includegraphics[width=0.45\textwidth]{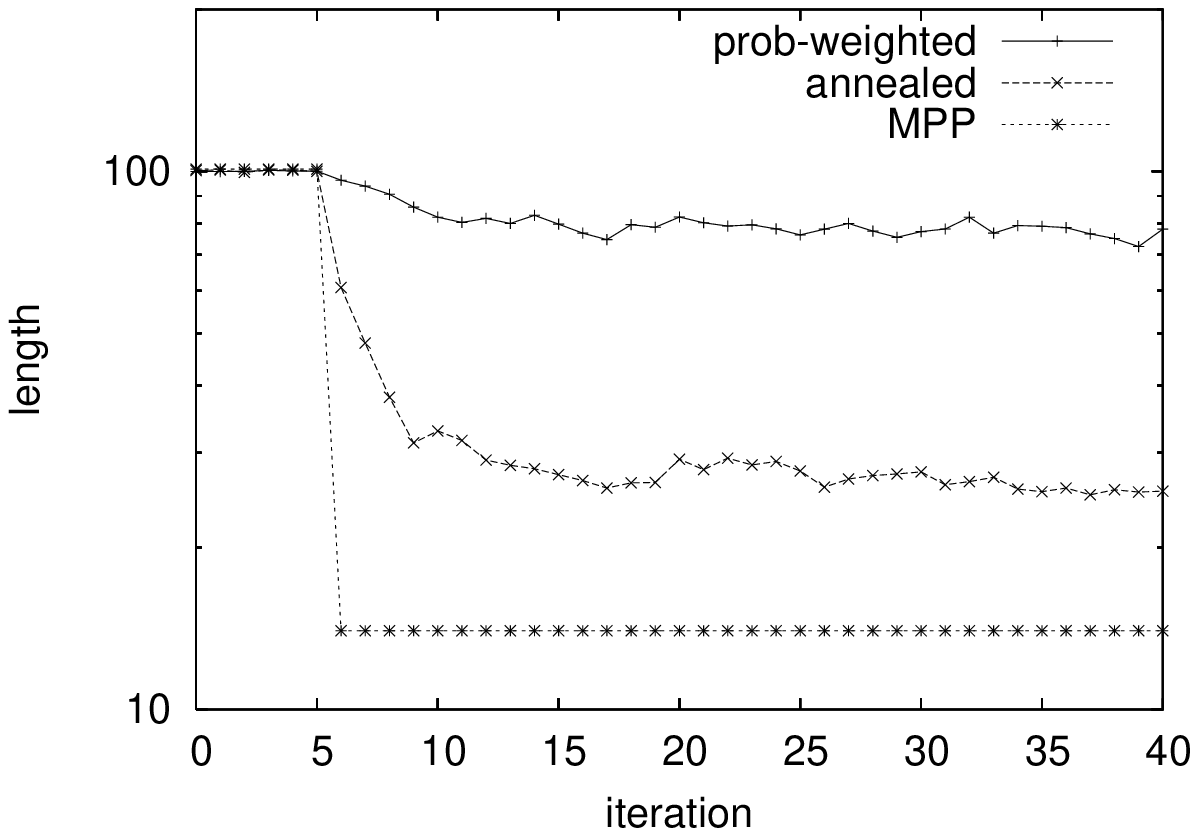}\hfill
  \includegraphics[width=0.45\textwidth]{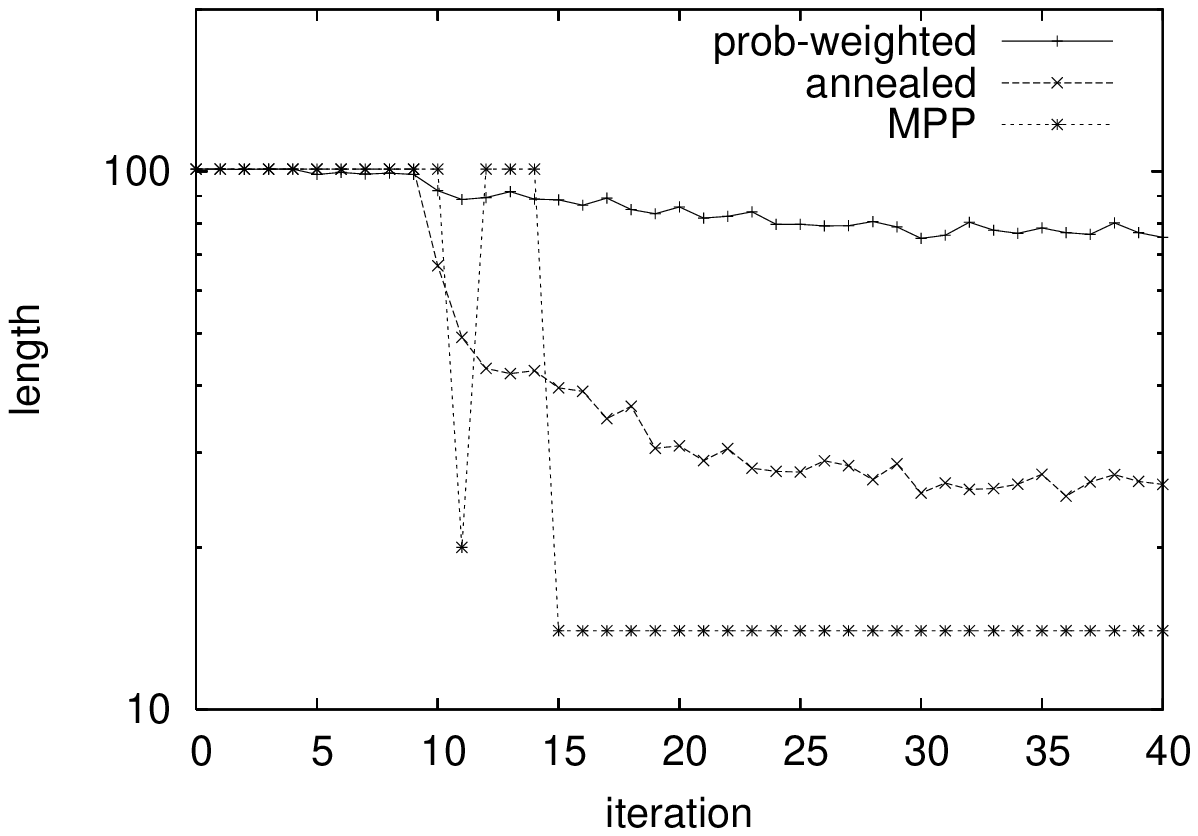}
\caption{%
  Plot of simulated path length versus GREP iteration of a small toy
  MDP maze for the probability weighted (PW) policy, annealed PW
  policy and MPP policy. The shortest path to goal is 14. Left:
  starting from initial uniform policy. Right: starting from initial
  random policy. }
\label{fig:reward}
\end{figure*}

We performed some numerical experiments using offline GREP. Our test
problem was a pure planning task in a $10\times10$ toy maze (see
Fig.~\ref{fig:maze}) where the probabilistic policy $\vec{P}$
represents the probability of taking a certain action at a certain
maze position. The same figure also shows typical solutions for the
quantities $\vec{z}$ and $\vec{q}$, i.e. the expected occupancy and
expected reward respectively (for certain $\vec{P}$).

After each GREP iteration, i.e. after each gradient calculation and
$\vec{P}$ update, we checked the obtained policy by running 20
simulations using the current value of $\vec{P}$. The probability
weighted (PW) policy selects action $k$ at state $i$ proportional to
$P_{ik}$, while the annealed PW policy uses an annealing factor of
$T=4$; we also simulated the MPP solution. Figure~\ref{fig:reward}
shows the average simulated path length versus GREP iteration of the
PW, the annealed PW policy and the derived MPP policy. In the left
plot the initial policy $\vec{P}$ was taken uniform. The right plot in
the same figure shows the simulated path lengths from a random policy;
also here the MPP finds the optimal solution but slightly later.

We see from the figure that in both cases the probability-weighted
(PW) policy is improving during the GREP iterations. However, the
convergence is very slow which shows the severe non-linearity of the
problem. The annealed PW does perform better than PW. Finally, we see
that MPP finds the optimal solution quickly within a few iterations.
Using Dijkstra's method, we confirmed that the found MPP policy was in
agreement with the global shortest path solution.

\subsection*{Online GREP}
The account below desribe an idea to use GREP when the environment is
not known beforehand\footnote{At the time of writing, we have not
  implemented this idea yet}. The steps actually interleave ``Kalman
filter''-like estimation of the unknown environment transition
probabilities with the explicit planning of GREP. In fact, it also
includes a step to estimate a possibly unknown (linear) sensor
mapping. Apart from the policy matrix $\vec{P}$, we need to estimate
also the (environment) transition probabilities $T_{kji}$ and possibly
sensor matrix $\vec{B}$. We can optimize for all parameters by
iteratively ascending to their conditional mode. The conditional
maximizing steps are easy:
\begin{enumerate} 
\item {\em Plan ahead:} Compute the policy gradient $\vec{G}$ in
  Eq.~\ref{eq:gradient} and improve current policy
  \begin{equation}
    \vec{P} \leftarrow \vec{P} + \alpha \vec{G}
    \label{eq:t-update}
  \end{equation}
  where $\alpha$ is a suitable step size parameter; for efficiency we
  can also perform a linesearch. After, we need to renormalize the
  columns of $\vec{P}$. See note below on policy regularization.

\item {\em Select action:} Given state estimate $\vec{s}_t$, draw an
  action $k$ from the policy according to:
  \[ k \sim \vec{P}_{\!t} \, \vec{s}_t. \]
  and receive reward $R$ and estimate new state $\vec{s}_{t+1}$.

\item {\em Estimate state:} Observe $\vec{y}_{t+1}$ and estimate
  $\vec{s}_{t+1}$ using
  \begin{equation}
    p(\vec{s}_{t+1} | \vec{y}_{t+1}, \vec{s}_{t}, k ) 
    \; \propto \;
    p(\vec{y}_{t+1} | \vec{s}_{t+1}) \, 
    p(\vec{s}_{t+1} | \vec{F}_k, \vec{s}_{t} ).
  \end{equation}
  Assuming Gaussian noise for observations and state estimates we
  obtain:
  \begin{equation}
    \vec{s}_{t+1} = \frac{
    \vec{H}_s \vec{F}_k, \vec{s}_{t} +
    \vec{B}^{-1} \vec{H}_y \vec{y}_{t+1} }
  { \vec{H}_s + \vec{B}^{-1} \vec{H}_y \vec{B}^{-T} }
  \end{equation}
  where $\vec{B}$ is the \emph{sensor matrix} that maps internal state
  $\vec{s}$ to observations $\vec{y}$. Matrices $\vec{H}_s$ and
  $\vec{H}_y$ are the inverse covariance (or so called
  \emph{precisions}) of state $\vec{s}$ and observation $\vec{y}$
  respectively.
  
\item {\em Estimate sensors:} In case also sensor matrix $\vec{B}$ is
  unknown, we have to perform an additional estimation step for
  $\vec{B}$. This is common step in the standard Kalman formulation.

\item {\em Estimate environment:} Given action $k$ and reward $R$, we
  update the reward vector
  \begin{equation}
    r_j \leftarrow R
  \end{equation}
  and the environment transition probabilities
  \begin{equation}
   \dd \vec{F}_k \;\propto\; (\vec{s}_{t+1} - \vec{F}_k \vec{s}_t )
   \vec{s}_t^T  (\vec{s}_t \vec{s}_t^T)^{-1},
 \end{equation}
 or if $(\vec{s}_t \vec{s}_t^T)^{-1}$ does not exist we can use
 \begin{equation}
   \dd \vec{F}_k \;\propto\; (\vec{s}_{t+1} - \vec{F}_k \vec{s}_t ) 
   (\vec{s}_t^T \vec{s}_t)^{-1} \vec{s}_t^T.  
 \end{equation}
  where $\vec{s}_t = j$, and reestimate the environment transition
  probabilities
  \begin{equation}
    \vec{T}_{k} \leftarrow
    \vec{T}_{k} + (\vec{s}_{t+1} - \vec{T}_k \vec{s}_t ) \vec{s}_t^T.  
  \end{equation}
  After the update, one should set entries in $\vec{F}'_k$ that
  corresponds to physically impossible transitions to zero.
  After, we need to renormalize the columns of $\vec{T}_k$. It is
  important to note that given $\vec{s}_{t}$ and $\vec{s}_{t+1}$
  transition matrix $\vec{T}_k$ is conditionally independent of the
  policy $\vec{P}$. That is to say, we can obtain an accurate model of
  the environment using, e.g., just a random walk.
\item Repeat 1.
\end{enumerate}

To draw a picture of what is happening. In the planning stage, based
on the current (and maybe not accurate) environment model, the agent
tries to improve its current policy by planning ahead using the
gradient in Eq.~\ref{eq:gradient-p}. Remember that the gradient
involves simulating paths from the current state and adjoint paths
from the goal. In the action stage the agent samples an action from
its policy. Then the agent senses the new state and updates its
environment model using this new information. Notice that policy
improvement is not done ``backwards'' as traditionally is done in DP
methods but ``forward'' by planning ahead.

\section{Conclusions}

\subsubsection*{Future topics}
We have tacitly assumed that $\vec{z}$ and $\vec{q}$ are computed
using the same discount factor $\gamma$. However, we could introduce
separate parameters $\gamma_z$ and $\gamma_q$ which effectively
assigns a different ``forward time window'' for $\vec{z}$ and a
``backward time window'' for $\vec{q}$. In fact when $\gamma_z
\rightarrow 0$ we have a ``one-step-look-ahead''. Alternatively, in
the limit of $\gamma_q \rightarrow 0$ we obtain a gradient for a
greedy policy that maximize only ``immediate reward''. How both
parameters affect GREP's performance is a topic for future research.

The above suggests that GREP can be viewed as a generalization to
``one-step-look-ahead'' policy improvement.  In fact, a
``one-step-look-ahead'' improvement rule using can be obtained for
$\gamma_z \rightarrow 0$ simply by taking $\vec{z}=\vec{s}_t$ in
Eq.~\ref{eq:gradient-p}.  Such an approach would be ``policy greedy''
in a sense that it updates the policy only locally. We expect GREP to
perform better because it updates the policy more globally; whether
this in fact improves GREP is also a remaining issue for future
research.

The interleaving of GREP with a Kalman-like estimation procedure of
the environment could handle a variety of interesting problems such as
planning in POMDP environments. 

We must mention that appropriate reparameterization of the stochastic
policy, e.g. using a Boltzman distribution, could improve the
convergence. We have not pursued this further.

\subsubsection*{Summary}
We have introduced a learning method called ``gradient-based
reinforcement planning'' (GREP). GREP needs a model of the
environment and plans ahead to improve its policy \emph{before} it
actually acts in the environment.  We have derived formulas for the
exact policy gradient. 

Numerical experiments suggest that the probabilistic policy indeed
converges to an optimal policy---but quite slowly. We found that (at
least in our toy example) the optimal solution can be found much
faster by annealing or simply by taking the most probable action at
each state.

Further work will be to incorporate GREP in online RL learning tasks
where the environment parameters, i.e. transition probabilities
$T_{kji}$, are unknown and have to be learned. While an analytical
solution for $\vec{q}$ and $\vec{z}$ are only viable for small problem
sizes, for larger problems we probably need to investigate Monte Carlo
or DP methods.%

{\small 

\begin{thebibliography}{}

\bibitem[Baird, 1998][Baird][1998]{baird98gradient}
Baird, L.~C. (1998).
\newblock Gradient descent for general reinforcement learning.
\newblock {\em Advances in Neural Information Processing Systems}.
\newblock {MIT} Press.

\bibitem[Baxter \& Bartlett, 1999][Baxter and Bartlett][1999]{baxter99direct}
Baxter, J., \& Bartlett, P. (1999).
\newblock {\em Direct gradient-based reinforcement learning: I. gradient
  estimation algorithms} (Technical Report).
\newblock Research School of Information Sciences and Engineering, Australian
  National University.

\bibitem[Difilippo et~al.\/, 1996][Difilippo et~al.\/][1996]{difilippo}
Difilippo, F.~C., Goldstein, M., Worley, B.~A., \& Ryman, J.~C. (1996).
\newblock Adjoint {M}onte {C}arlo methods for radiotherapy treatment planning.
\newblock {\em Trans. Am. Nucl. Soc.}, {\em 74}, 14--16.

\bibitem[Ng et~al.\/, 1999][Ng et~al.\/][1999]{ng99policy}
Ng, A., Parr, R., \& Koller, D. (1999).
\newblock Policy search via density estimation.
\newblock {\em Advances in Neural Information Processing Systems}.
\newblock {MIT} Press.

\bibitem[Schmidhuber, 1990][Schmidhuber][1990]{Schmidhuber:90sandiego}
Schmidhuber, J. (1990).
\newblock An on-line algorithm for dynamic reinforcement learning and planning
  in reactive environments.
\newblock {\em Proc. IEEE/INNS International Joint Conference on Neural
  Networks, San Diego} (pp.\/ 253--258).

\bibitem[Sutton \& Barto, 1998][Sutton and Barto][1998]{suttonbarto}
Sutton, R.~S., \& Barto, A.~G. (1998).
\newblock {\em Reinforcement learning. {A}n introduction}.
\newblock {MIT} Press, Cambridge.

\bibitem[Sutton et~al.\/, 2000][Sutton et~al.\/][2000]{sutton00policy}
Sutton, R.~S., McAllester, D., Singh, S., \& Mansour, Y. (2000).
\newblock Policy gradient methods for reinforcement learning with function
  approximation.
\newblock {\em Advances in Neural Information Processing Systems}.
\newblock {MIT} Press.

\bibitem[Williams, 1992][Williams][1992]{williams92simple}
Williams, R.~J. (1992).
\newblock Simple statistical gradient-following algorithms for connectionist
  reinforcement learning.
\newblock {\em Machine Learning}, {\em 8}, 229--256.

\end{thebibliography}

}


\section*{Appendix A: Implicit policies}
In deterministic environments where the state-action pair $(s_i,a_m)$
uniquely leads to a state $s_j$, i.e. $T_{kji} = \delta_{km}
(k=1,...,K)$ the projection $\vec{F}$ is solely determined
by the policy $\vec{z}$, and \emph{vice versa}. We refer to this as
the case of \emph{implicit policy} because the policy is implicitly
implied in the induced transition probability $T_{ji}$.

In such environments we can suffice to solve for $\vec{F}$ directly
and omit parameterization through $\vec{z}$. From
Eq.~\ref{eq:projection} we see that
\begin{equation}
  P_{im} = T_{ji}
\end{equation}
and using a similar derivation as we have done for $\partial H /
\partial P_{ik}$, it can be shown that the gradient of $H$ with
respect to $\vec{F}$ is given by
\begin{equation}
  \vec{G} = \vec{r} \vec{z}^T.
\label{eq:rank-one}
\end{equation}
An important point must be mentioned. In most cases many elements
$T_{ji}$ are zero, representing an absent transition between $S_i$ and
$S_j$. Naively updating $\vec{F}$ using the full gradient $\vec{G}$
would incur complete fill-in of $\vec{F}$ which is in most cases not
desirable or even physically incorrect. Therefore, one must check the
gradient each time and set impossible transition probabilities to
zero. We will refer to this ``heuristically corrected'' gradient as
$\widetilde{\vec{G}}$. Also, after each update, we have to renormalize
the columns of $\vec{F}$. The rank-one update in Eq.~\ref{eq:rank-one}
is interesting because it provides an efficient means of calculating
the inverse in Eq.~\ref{eq:occupancy}.


\section*{Appendix B: Monte Carlo gradient sampling}

In our example, we calculated $\vec{z}$ and $\vec{q}$ in
Eq.\ref{eq:occupancy} by linear programming. For large state spaces
the matrix inversion quickly becomes too computationally intensive and
probably traditonal dynamic programming based methods would be more
efficient.

Instead, we investigated to use Monte Carlo (MC) simulation. We use
\emph{forward sampling} to approximate the expected state occupancies
in $\vec{z}$ and use, so-called, \emph{adjoint Monte Carlo
  sampling}~\cite{difilippo} to estimate the adjoint reward $\vec{q}$.
Adjoint MC simulation of is far more efficient than would
we have estimated each $q_i$ by a separate MC run.%
\footnote{With one MC run we mean performing, e.g., 10000 trials from
  a fixed state.} %
By performing the simulation backward from $\vec{r}$, we obtain all
values of $\vec{q}$ using only a single MC run.

\begin{figure} \centering
  \includegraphics[height=0.28\textwidth]{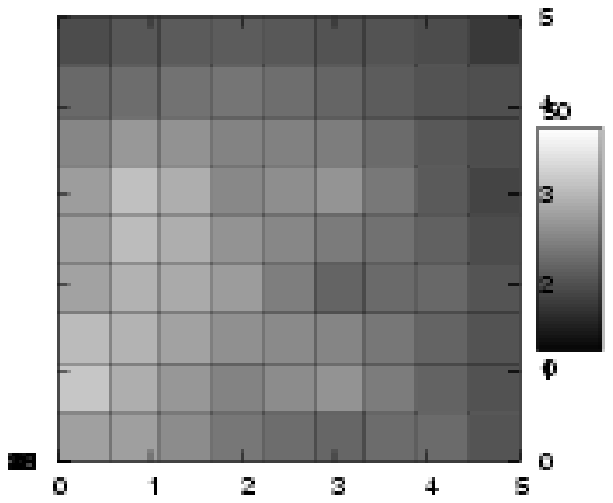} \hfill
  \includegraphics[height=0.28\textwidth]{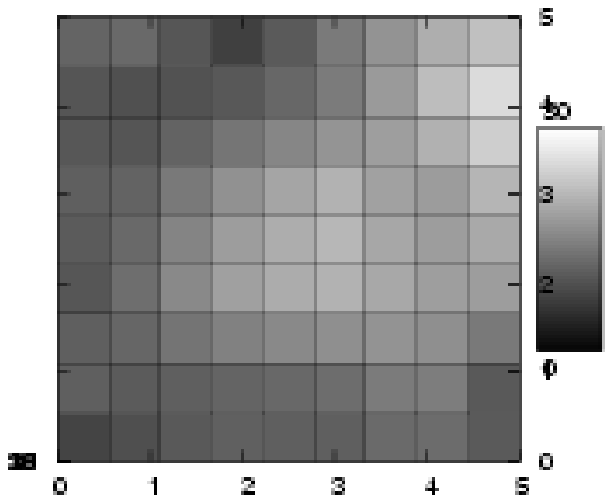} \hfill
  \includegraphics[height=0.28\textwidth]{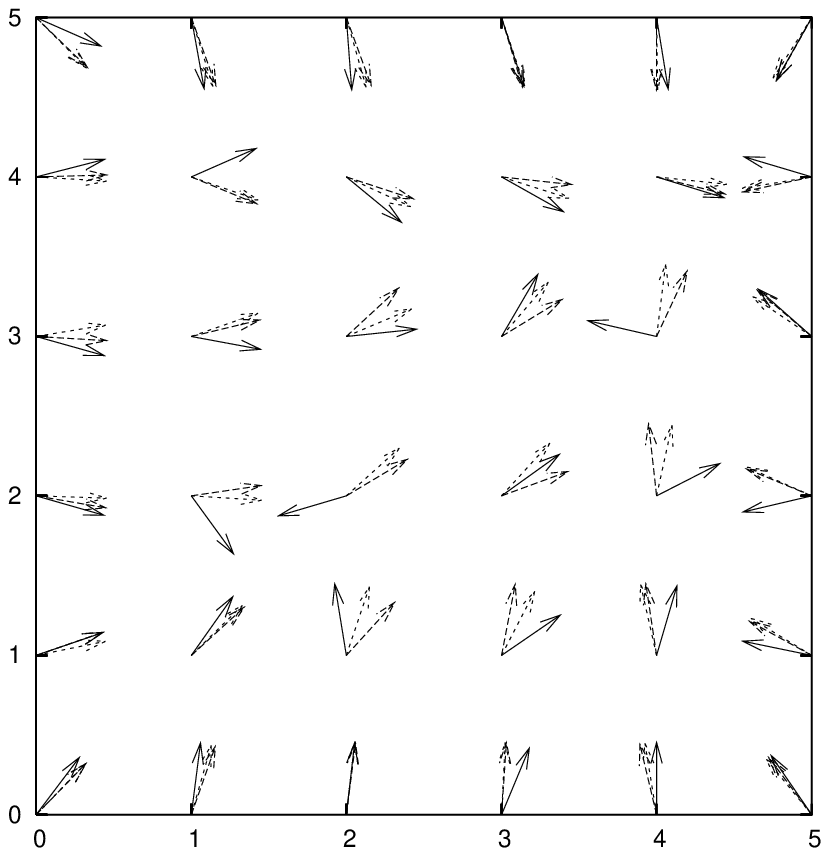} \\
  \hspace{24mm} (a) \hfill (b) \hfill (c) \hspace{20mm}
\caption{ \it MC calculations of a $6\times6$ toy maze. The agent is
  at (0,1) and targets (5,4).  (a) Expected occupancy, (b) adjoint
  probability, and (c) normalized policy gradient for $n=4$, $n=40$,
  $n=200$. Each vector is computed as $\vec{v} = \sum_k P_{ik}
  \vec{e}_k$ where $\vec{e}_k$ is the unit vector along the state
  change induced by action $a_k$.}
\label{fig:mc-gradientplots}
\end{figure}

Fig.~\ref{fig:mc-gradientplots} shows the MC approximations of
$\vec{z}$ and $\vec{q}$. On the right of the same figure, we have
plotted the computed policy-gradient based on MC estimates using a
minimum number of $n=\{20, 40, 200\}$ samples. To compare them with
the exact gradient, we calculated the exact values of $\vec{z}$ and
$\vec{q}$ by inverting the linear system.  For larger number of
samples, the gradient vector do indeed point more strongly towards the
goal.

An important feature of general Monte Carlo methods is that they
automatically concentrate their sampling to the important regions of
the parameter space ---mostly proportional to the posterior or the
likelihood. For our purpose of sampling the gradient, to even more
concentrate the sampling density towards the regions of large gradient
values, we have tried to apply \emph{annealing}. To sample from a
density $p(\theta)$ we may sample from the annealed function
$p_\gamma(\theta) = p(\theta)^\gamma / \int p(\theta)^\gamma d\theta$
and reweight each sample with its importance weight
$1/p_\gamma(\theta)$. For $\gamma \rightarrow \infty$, the set of
samples converges to the maximum probable gradient.
 
In conclusion, our approach of separately estimating $\vec{q}$ and
$\vec{z}$ using MC and \emph{then} (elementwise) multiply their
solutions, doesn't really brought clear advantages. If we could sample
from the joint distribution $q_i z_i$ (i.e. elementwise product) then
MC would clearly turn out to be a very efficient method.

\end{document}